\documentclass[letterpaper, 10 pt, conference]{ieeeconf}  % Comment this line out if you need a4paper

\IEEEoverridecommandlockouts                             % This command is only needed if 
\usepackage{bm}
\usepackage{amssymb}
\usepackage{threeparttable}
\usepackage{wrapfig}
\usepackage{amsmath}
\usepackage{wrapfig}
\usepackage{lipsum}
\usepackage{siunitx}
\usepackage{booktabs}
\usepackage{glossaries}
\usepackage{hyperref}
\usepackage{graphicx}
\usepackage{url}
\usepackage{cleveref}
\usepackage{cite}
\usepackage{xcolor}
\usepackage{times} % assumes new font selection scheme installed

\title{\small [THIS IS A PREPRINT] The paper will be published at IROS23\\
\LARGE \bf Anytime, Anywhere: Human Arm Pose from Smartwatch Data for Ubiquitous Robot Control and Teleoperation}

\author{Fabian C Weigend, Shubham Sonawani, Michael Drolet and Heni Ben Amor %\vspace{-0.2cm}% <-this % stops a space
\thanks{All Authors are with the School of Computing and Augmented Intelligence, Arizona State University \texttt{$\lbrace$fweigend, sdsonawa, mdrolet, hbenamor$\rbrace$@asu.edu}%
}
}

\begin{document}
\newacronym{IMU}{IMU}{Inertial Measurement Units}
\newacronym{PPG}{PPG}{Photoplethysmography}
\newacronym{LSTM}{LSTM}{Long Short-Term Memory}
\newacronym{MAE}{MAE}{Mean Absolute Error}
\newacronym{6DRR}{\text{6DRR}}{6-dimensional rotation representation}
\newacronym{MC}{MC}{Monte Carlo}
\newacronym{RMSE}{RMSE}{root mean squared error}
\newacronym{3D}{3D}{3-dimensional}
\newacronym{SELU}{SELU}{scaled exponential linear units}
\newacronym{GAIL}{GAIL}{Generative Adversarial Imitation Learning}
\newacronym{gyro}{\ensuremath{\bm{\phi}}}{gyroscope measurements}
\newacronym{grav}{\ensuremath{\bm{\gamma}}}{gravity sensor}
\newacronym{lacc}{\ensuremath{\bm{\alpha}}}{linear acceleration sensor}
\newacronym{racc}{\ensuremath{\bm{\alpha}_\mathrm{raw}}}{raw acceleration}
\newacronym{swrot}{\ensuremath{\bm{\theta}}}{virtual rotation vector sensor}
\newacronym{swrot_calib}{\ensuremath{\bm{\theta}_c}}{calibration forward-facing direction}
\newacronym{swrot_r}{\ensuremath{\bm{\theta}^r}}{relative smartwatch rotation}
\newacronym{pres}{\ensuremath{\rho}}{atmospheric pressure sensor}
\newacronym{pres_chest}{\ensuremath{\rho_\mathrm{c}}}{atmospheric pressure at chest height}
\newacronym{rpres}{\ensuremath{\rho^r}}{relative atmospheric pressure}
\newacronym{pos_wrist}{\ensuremath{\textbf{p}_\mathrm{w}}}{wrist position}
\newacronym{pos_elbow}{\ensuremath{\textbf{p}_\mathrm{e}}}{elbow position}
\newacronym{pos_shoul}{\ensuremath{\textbf{p}_\mathrm{s}}}{shoulder position}
\newacronym{rot_hip}{\ensuremath{\textbf{q}_\mathrm{h}}}{hip rotation}
\newacronym{rot_larm}{\ensuremath{\textbf{q}_\mathrm{l}}}{lower arm rotation}
\newacronym{rot_uarm}{\ensuremath{\textbf{q}_\mathrm{u}}}{upper arm rotation}
\newacronym{rot_larm_r}{\ensuremath{\textbf{q}^r_\mathrm{l}}}{relative lower arm rotation}
\newacronym{rot_uarm_r}{\ensuremath{\textbf{q}^r_\mathrm{u}}}{relative upper arm rotation}
\newacronym{larm_len}{\ensuremath{l_l}}{lower arm length}
\newacronym{uarm_len}{\ensuremath{l_u}}{upper arm length}

\maketitle
\thispagestyle{empty}
\pagestyle{empty}

% A smartwatch is a consumer-grade low-cost device that users are familiar with. 
%%%%%%%%%%%%%%%%%%%%%%%%%%%%%%%%%%%%%%%%%%%%%%%%%%%%%%%%%%%%%%%%%%%%%%%%%%%%%%%%
\begin{abstract}
    This work devises an optimized machine learning approach for human arm pose estimation from a single smartwatch. Our approach results in a distribution of possible wrist and elbow positions, which allows for a measure of uncertainty and the detection of multiple possible arm posture solutions, i.e., multimodal pose distributions. Combining estimated arm postures with speech recognition, we turn the smartwatch into a ubiquitous, low-cost and versatile robot control interface. We demonstrate in two use-cases that this intuitive control interface enables users to swiftly intervene in robot behavior, to temporarily adjust their goal, or to train completely new control policies by imitation. Extensive experiments show that the approach results in a 40\% reduction in prediction error over the current state-of-the-art and achieves a mean error of 2.56\,cm for wrist and elbow positions. The code is available at \url{https://github.com/wearable-motion-capture}.
\end{abstract}
%%%%%%%%%%%%%%%%%%%%%%%%%%%%%%%%%%%%%%%%%%%%%%%%%%%%%%%%%%%%%%%%%%%%%%%%%%%%%%%%

\section{INTRODUCTION}

The relationship between humans and robots is a central question of artificial intelligence and robotics. As robots become increasingly capable, there is growing interest for human-robot collaboration in various domains, such as healthcare, manufacturing, and daily activities. Many scenarios in these fields envision humans to teleoperate, assist, or teach a robot counterpart. For example, a human expert may demonstrate to a robot how to perform a new task or how to manipulate a new object. Such scenarios, however, require intuitive and robust interfaces for capturing human body motion. 

To date, motion capture cameras are the gold standard in capturing human motion \cite{aurand_accuracy_2017,nagymate_application_2018,vilas-boas_full-body_2019}. A setup of multiple cameras can provide a high-fidelity recording of body postures and positions over time. However, motion capture requires an expensive and stationary setup. Easier consumer-grade hardware, e.g., Microsoft Kinect, provides only low-fidelity approximations of the body posture and is heavily affected by line-of-sight issues and a limited field-of-view \cite{vilas-boas_full-body_2019}. Alternative motion capture approaches are based on \gls{IMU} and allow tracking without line-of-sight issues. However, they typically require wearing two or more \gls{IMU}s on different limbs, e.g., strapped around lower arm and upper arm or as a special suit \cite{joukov_human_2017,li_mobile_2020,ates_force_2022}. Even though research has investigated human arm posture estimations from a single \gls{IMU}, the authors in \cite{shen_i_2016,wei_real-time_2021} reported prediction accuracy is of low fidelity.

%Postures cannot be properly detected when one or more limbs of the user are occluded by another body part, person, or object. 

\begin{figure}[htbp]
\centerline{\includegraphics[width=0.46\textwidth]{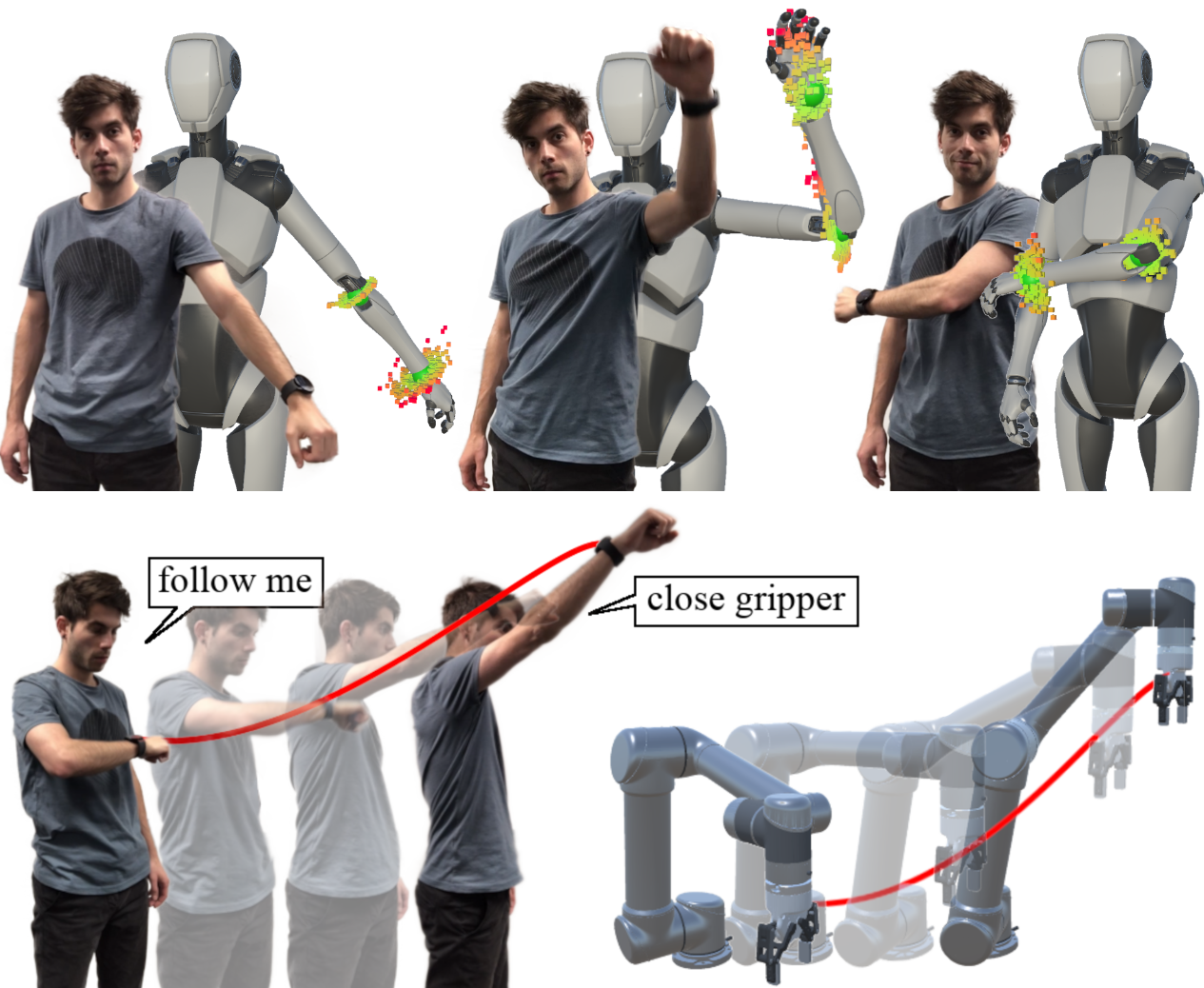}}
\caption{\textbf{Top:} The avatar shows predicted elbow and wrist positions from smartwatch sensor data. Our approach results in a distribution of solutions. The mean of a distribution is depicted as a green sphere %and determines the final predicted position of wrist and elbow respectively. 
All individual predictions of a distribution are depicted as small cubes, colored according to their proximity to the mean. \textbf{Bottom:} We also stream microphone data to utilize speech recognition. This combination offers a versatile interface to interact with and to control robots anytime and anywhere.}
\label{fig:teaser}
\vspace{-\baselineskip}
\end{figure}

In this paper, we devise a machine learning approach to increase the accuracy for predicting human arm poses from the single \gls{IMU} of a smartwatch. As observable in the top in \Cref{fig:teaser}, our approach results in a distribution of predicted postures, which allows to estimate a measure of uncertainty and provides a range of possible solutions to pick from. 

By combining the increased accuracy of our approach with speech recognition, we turn the smartwatch into an ubiquitous robot control interface. Smartwatches are widely recognized as common consumer-grade devices that users are already familiar with~\cite{yang_survey_2022}. Without the need for a complicated setup, a human expert can engage with the robot at any time and anywhere. As depicted in \Cref{fig:teaser}, they may move the robot to a new target and issue commands via speech recognition. We summarize our contributions as follows:

\begin{itemize}
    \item We present a machine learning approach for real-time estimation of upper and lower arm postures from a single smartwatch. 
    \item Our approach results in a distribution of possible arm postures, which opens up opportunities for selecting optimal solutions. %that, for example, maximize safety constraints on a teleoperated robot.
    \item We identify solutions to calibration, data representation, and network design that yield higher accuracy than previously reported results in the literature. 
    \item We combine human arm posture estimations with speech recognition and present two real-robot examples that highlight the advantages of our smartwatch approach for robotics.
\end{itemize}

\section{RELATED WORK}
% yamane_kinematic_2020
Tracking one or multiple parts of the human body is an essential step in approaches to robot control. For example, techniques for teleoperation build upon the accurate detection of human body pose~\cite{darvish_teleoperation_2023}. In a similar vein, imitation learning~\cite{osa_algorithmic_2018} or programming-by-demonstration~(PbD)~\cite{billard_robot_2008} requires a human expert to provide one or more demonstrations of target motions. These are distilled into a policy that generalizes the observed behavior to new situations. Traditionally, a large number of works for PbD have relied on costly motion capture setups for recording high-fidelity data~\cite{ijspeert_movement_2002, ott_motion_2008, hasenclever_comic_2020}. Other approaches try to strike a balance between the cost of data collection and the fidelity by leveraging Inertial Measurement Units (IMUs) or camera-based setups. For example, the works in~\cite{ates_force_2022} use multiple IMUs attached to different parts of the body to transfer human motions onto a robot. However, approaches based on multiple IMUs require a careful placement of sensors on the human body along with a (potentially time-consuming) calibration process.
% -- providing data to a robot cannot be performed \emph{impromptu}.

More recently, consumer-grade hardware for virtual and augmented reality (VR/AR) is becoming an alternative for motion tracking in robotics~\cite{rakita_motion_2017, dyrstad_teaching_2018, hirschmanner_virtual_2019}. For example, the work in~\cite{rakita_motion_2017} uses a HTC Vive VR system for robot teleoperation in a manipulation task. HTC Vive controller estimate their positions from infrared signals from so-called base stations, which have to be carefully placed and calibrated. In a similar vein, the work in~\cite{delpreto_helping_2020} uses an Oculus Quest device for upper body tracking. %Despite the low cost, VR devices and their respective joystick controllers can provide millimeter accuracy. 
However, Oculus controllers are tracked via cameras within the VR headset \cite{yim_wfh-vr_2022}. Headsets can cause ergonomic discomfort and reduce the situational awareness.

%The main objective is provide an intuitive interface for providing instructions to a robot seamlessly. 
Wearable devices like a smartwatch can only provide comparably low-fidelity position data. Instead, they offer a combination of low-cost, ease-of-use and a broad range of additional sensors, e.g., magnetometer, atmospheric pressure sensor, microphone or \gls{PPG} sensor~\cite{yang_survey_2022}. For example, these on-body sensors enable advances in emotion sensing~\cite{yang_survey_2022}. In robot control, smartwatches are mostly used to control robots with roll, pitch, and yaw estimates from \gls{IMU} and magnetometer~\cite{villani_interacting_2017}. Research has also investigated methods for human pose estimations from smartwatch data \cite{shen_i_2016,wei_real-time_2021}, however, these are of low precision and have mostly been intended for recreational purposes or physical therapy\cite{wei_real-time_2021}. To open up more opportunities to utilize the advantages of wearable devices in robot control, we propose a solution to improve real-time arm pose estimations in such settings. 

\section{METHODOLOGY}

In this work, we address the problem of estimating human arm poses from a single smartwatch. We cast the process as a supervised learning task, in which postural information is predicted from a set of multimodal sensors. A challenging aspect is the inherent one-to-many mapping imposed by redundant human kinematics. Readings obtained from smartwatch sensors may not correspond to a single arm movement or position, but rather, can indicate various possibilities. Another challenge emerges from natural variability in the sensor data. Sensor readings for pressure and orientation need to be adjusted before usage. In the following section, we discuss how to train deep learning models that are particularly well-suited to the requirements of the task. %We discuss a variety of solutions with regards to data collection, processing, representation and learning. 
%The combination of multiple insights and innovations allows us to devise an optimized predictive model architecture.

%For dealing with data variability, we suggest a straightforward smartwatch calibration procedure. With regards to the one-to-many problem, we leverage Monte Carlo (MC) dropout predictions to generate multimodal probability distributions from a single neural network. As a result, generated predictions include an estimate of the degree-of-uncertainty of the underlying prediction.

\subsection{Data Collection}
\label{subsec:data_collection}

We collect motion capture data as ground truth prediction targets and match these with recorded smartwatch sensor measurements. To this end, we develop a Wear OS app to record and stream sensor measurements. The app is tested on a Samsung Galaxy Watch 5. It records data from a set of multimodal sensors. These include \mbox{\gls{gyro}} with $\gls{gyro} \in \mathbb{R}^3$, which represent the angular velocities. Further, it records measurements of the \mbox{\gls{grav}} and \mbox{\gls{lacc}} with $\gls{grav},\gls{lacc} \in \mathbb{R}^3$, which represent the acceleration with respect to the X, Y and Z axis. Linear acceleration is the \gls{racc} minus the gravity measurements such that $ \gls{lacc} = \gls{racc} - \gls{grav}.$
In addition, the app records the \mbox{\gls{swrot}}, which is provided by Wear OS. The rotation vector sensor estimates the global smartwatch rotation from the magnetometer, accelerometer and gyroscope as a quaternion, thus, $\gls{swrot} \in \mathbb{R}^4$. Together with the reading from the \gls{pres} with $\gls{pres} \in \mathbb{R}$, one observation $s$ from the smartwatch consists of the following values $\bm{s} = [
    \gls{swrot}, 
    \gls{lacc},
    \gls{grav},
    \gls{gyro},
    \gls{pres}
    ]^\top$, 
with $\bm{s} \in \mathbb{R}^{14}$. In addition, the app also streams the microphone data, which we use for speech recognition. However, because we do not utilize microphone data for arm pose estimations, it is not included in $\bm{s}$.

\begin{figure}[t]
\centerline{\includegraphics[width=0.5\textwidth]{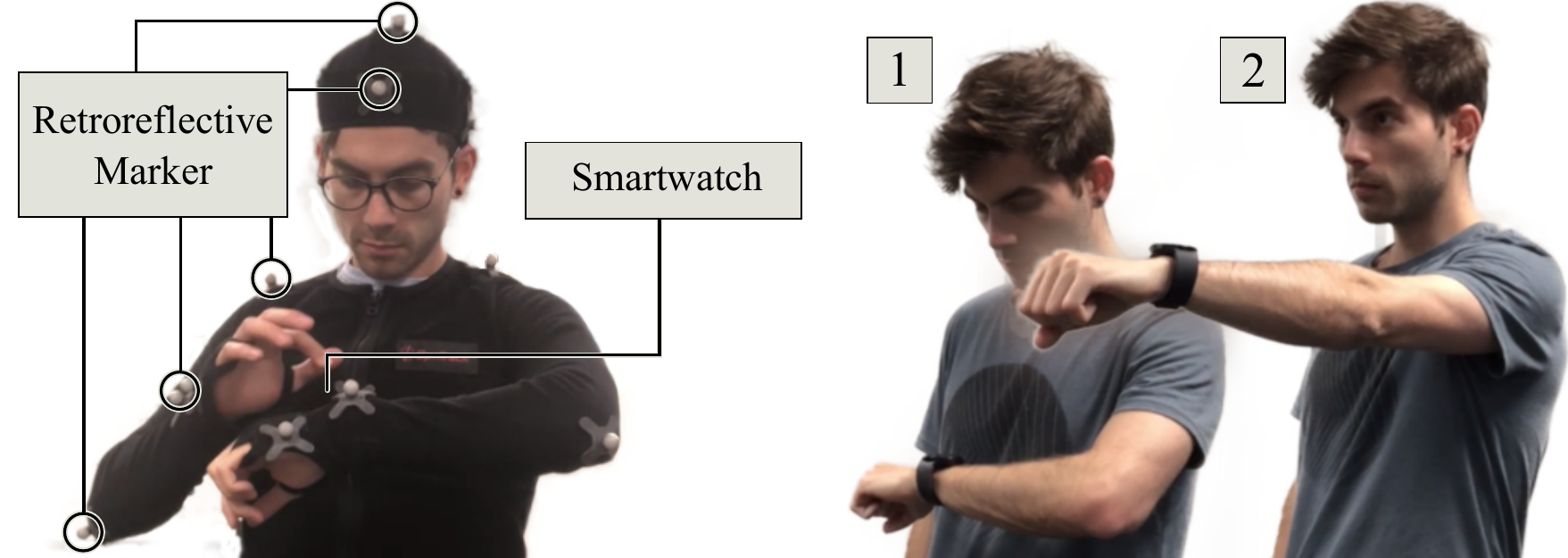}}
\caption{\textbf{Left:} We collected ground truth data with an optical motion capture system and a 25-marker upper body suit. \textbf{Right:} Our two-step calibration process. First, the user holds the watch at chest height to estimate relative atmospheric pressure. Then, the user stretches the arm forward for an estimate of body orientation.}
\label{fig:optical_marker_and_calibration_pictures}
\vspace{-0.05in}
\end{figure}

As ground truth, we collect upper-body motion capture data. We use the research-grade optical motion capture system OptiTrack \cite{nagymate_application_2018}. The motion capture environment features 12 cameras. %, which are calibrated before data collection. 
We recorded data from 6 participants, who wore a 25-marker-upper-body suit along with the smartwatch on their left wrist (See \Cref{fig:optical_marker_and_calibration_pictures}). We collect the \mbox{\gls{rot_hip}}, \mbox{\gls{rot_larm}} and \mbox{\gls{rot_uarm}} as quaternions. We further store the \mbox{\gls{larm_len}} and \mbox{\gls{uarm_len}} of the participant to estimate wrist and elbow positions from recorded rotations. Therefore, a motion capture ground truth observation $\bm{g}$ contains \mbox{$\bm{g} = [\gls{rot_hip}, \gls{rot_larm}, \gls{rot_uarm}, \gls{larm_len}, \gls{uarm_len}]^\top$}, with $\bm{g} \in \mathbb{R}^{14}$.

Once the motion capture system and our smartwatch app started recording, participants were instructed to keep their chest and hip stationary while moving their left arm in any possible way. The smartwatch recorded at around $\sim50$\,Hz which resulted in a set of 381\,535 observations. The motion capture system recorded at $\sim120$\,Hz which resulted in a set of 926\,164 motion capture observations. Data collection was conducted in accordance with Arizona State University~(ASU) guidelines. Written informed consent was obtained under and approved by the institutional review board~(IRB) of ASU under the ID~STUDY00017558.

\subsection{Data Processing}

Recorded smartwatch and motion capture data requires alignment and preprocessing since (a) motion capture data was recorded at a higher frequency, and (b) the data was collected in distinct coordinate systems. This subsection defines steps to merge smartwatch observations with our ground truth data. Additionally, we present a calibration procedure to further enhance correlations within the data and aid the training of predictive models for arm posture.

\begin{figure}[t]
\centerline{\includegraphics[width=0.5\textwidth]{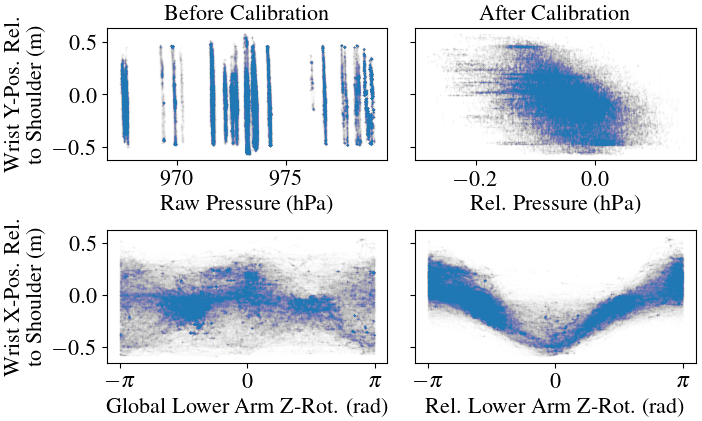}}
\caption{This figure depicts two examples for data before and after calibration. Each plot contains all of our 381\,535 data points.}
\label{fig:calibration_data}
\vspace{-\baselineskip}
\end{figure}

\textbf{Merging data sets} We retrospectively merged the set of collected smartwatch observations and the set of ground truth motion capture data by pairing observations according to their timestamps. Every smartwatch observation $\bm{s}$ was paired with the motion capture data observation $\bm{g}$ that was recorded closest in time. %As a result, we extend all 381\,535 observations of set $S$ with ground truth motion capture observations. 

\textbf{Calibrating atmospheric pressure} A critical smartwatch sensor provides measurements of the atmospheric pressure \gls{pres}. As depicted in the top left plot in \Cref{fig:calibration_data}, these measurements suffer from day-by-day variations due to changing weather conditions and temperature. Data that was collected in the same experiment or on the same day are recognizable as vertical lines when plotted against the corresponding \mbox{Y-position} (elevation) of the wrist. 

We propose a calibration procedure to remove the day-by-day variations and create a relative pressure measurement. %The first step of our calibration procedure is designed to calibrate atmospheric pressure measurements.
It is depicted on the right in \Cref{fig:optical_marker_and_calibration_pictures}. The user presses ``calibrate'' and holds the smartwatch at chest height. The watch records atmospheric pressure measurements for three seconds and then vibrates to signal that the step is completed. The average recorded pressure is saved as the \gls{pres_chest} used to estimate \gls{rpres} as $\gls{rpres} = \gls{pres} - \gls{pres_chest}$. 

The Kendall's Tau correlation coefficient between the \gls{pres} and wrist Y-position in the top left plot of \Cref{fig:calibration_data} is \mbox{-0.009}. In contrast, the Kendall's Tau correlation coefficient of \gls{rpres} and the wrist Y-position is -0.308, confirming that there is a higher correlation between the variables. We asked our participants to perform this calibration step before data collection and replaced the \gls{pres} measurement in $\bm{s}$ with \gls{rpres}.

\textbf{Calibrating rotation} Due to the kinematic structure underlying human anatomy, arm orientations are affected by the body orientation. However, no information about the body forward-facing direction is available from the smartwatch sensors. Although a universal solution is preferable, we introduce a constraint to overcome this hurdle: the forward-facing direction of the user must be constant and known. We explore future opportunities in this area in \Cref{sec:limt_and_future}, but for now, we will highlight the advantages of this imposed constraint in our approach.

We incorporate the constraint of a constant body forward-facing direction with the second step of our proposed two-step calibration procedure. The step is depicted under number two in \Cref{fig:optical_marker_and_calibration_pictures}. After completing the first step for the relative pressure measurement, the user stretches their arm forward. The watch records its rotation measurements for three seconds and saves their average as the \gls{swrot_calib}. This allows us to estimate the \gls{swrot_r} as the quaternion \mbox{
$\gls{swrot_r} = \gls{swrot_calib}^{-1} \gls{swrot}$}.

To transform our ground truth motion capture data into the same local coordinate system, we use the collected \gls{rot_hip} as the ground-truth forward-facing direction and estimate the \gls{rot_larm_r} and \gls{rot_uarm_r} as $\gls{rot_larm_r} = \gls{rot_hip}^{-1}\gls{rot_larm}$ and $\gls{rot_uarm_r} = \gls{rot_hip}^{-1}\gls{rot_uarm}$. Together with the saved lower and upper arm lengths, this information also allows us estimate wrist and elbow positions from these orientations in the same local coordinate system and relative to the shoulder.  

The example in the bottom plots of \Cref{fig:calibration_data} shows the benefit of using rotations relative to the forward-facing direction of the user. The rotation is denoted in Euler angles for easier interpretation. The body coordinate system in this example has the \mbox{Z-axis} tangential to the ground pointing forward and the \mbox{X-axis} along the right arm in T-pose. The Y-axis is orthogonal to the ground pointing upwards. This is a left-handed coordinate system. As observable in the bottom-right plot of \Cref{fig:calibration_data}, when the user extends their left arm wearing the smartwatch to the left, the lower arm \mbox{Z-rotation} from the T-pose is $0$ and the distance from wrist X-position to shoulder X-position is around -0.5\,m. In contrast, in the bottom-left plot, the global smartwatch rotation provides less information because users were not always facing the same direction during data collection. Thus, the relative rotation after our calibration allows to narrow down possible wrist positions from observed lower arm rotations. %For example, the left wrist will not be on the right of the shoulder (positive X) if the relative \mbox{Z-rotation} is 0 (left lower arm extended to the left). 

%In conclusion, we presented steps to merge collected smartwatch and ground truth motion capture data. We introduced the constraint that users have to maintain a constant body forward-facing direction. Further, we introduced a two-step calibration procedure which emphasizes and reveals correlations in our collected data. 

\subsection{Predictive Models}
\label{subsec:models}
%The proposed two-step calibration procedure is completed in seconds and allows us to approach a solution for the challenge of predicting arm posture from smartwatch data. 
Building upon presented data merging and calibration steps, we devise an optimized predictive model that benefits from this data preprocessing. To this goal, we investigate two distinct neural network architectures and four distinct representations of prediction targets. This allows us to compare and choose among a range of design choices which we present in the following. 

\textbf{Architectures and Inputs} We train two neural network architectures on two similar sets of inputs. The first architecture is a feedforward network, which receives as inputs $[\gls{rpres}, \gls{swrot_r}, \gls{lacc}, \gls{gyro}, \gls{grav}, \gls{larm_len}, \gls{uarm_len}]^\top$. The second architecture is an \gls{LSTM} network which receives the same input data with two additions: The data is stacked into a sequence of length 6 and it receives the time delta from each sequence step to the next.

\textbf{Prediction Targets} By human arm pose estimation from smartwatch data, more specifically, we refer to predicting the ground truth relative lower and upper arm rotations, i.e., \gls{rot_larm_r} and \gls{rot_uarm_r}, or predicting ground truth wrist and elbow positions which were estimated from these rotations. The naive way to predict wrist and elbow positions is to train a network to generate positions in Cartesian XYZ coordinates. However, since lower and upper arm lengths are constants, i.e. \gls{larm_len} and \gls{uarm_len}, we know that positions lie on a manifold, which allows to narrow down the search space. The elbow position has to lie on a sphere around the shoulder with radius \gls{uarm_len}. The wrist position has to lie on the manifold defined by spheres with a radius of \gls{larm_len} around all possible elbow positions \cite{shen_i_2016}. Therefore, as an alternative, we train our network architectures to predict upper and lower arm rotations and estimate positions from using known \gls{larm_len} and \gls{uarm_len}. Intuitively, polar coordinates come to mind as a suitable representation. When using \gls{uarm_len} as the radius, the position of the elbow relative to the shoulder is well-described by two angles. Further, rotations can be represented in quaternions.

However, these representation spaces do not have a continuous mapping to their rotation space, e.g., Euler angles jump from 359 to 0 degrees, which can cause complications during the training process due to discontinuity~\cite{zhou_continuity_2019}. A \gls{6DRR} has been proposed by \cite{zhou_continuity_2019}, with which the authors achieved promising results for training neural networks on a human pose inverse kinematics test. In the case of the \gls{3D} rotation group SO(3) in the \gls{3D} Euclidean space, their \gls{6DRR} space consists of the first two columns ($\textbf{a}_1$ and $\textbf{a}_2$) of the \gls{3D} rotation matrix. A mapping $g$ from rotation matrix to \gls{6DRR} is therefore:
\begin{equation}
g
\Biggl(
\;
\underbrace{
\begin{bmatrix}
| & | & | \\
\textbf{a}_1 & \textbf{a}_2 & \textbf{a}_3 \\
| & | & | \\
\end{bmatrix}
}_\text{Rotation Matrix} 
\;
\Biggr)
=
\underbrace{
\begin{bmatrix}
| & |  \\
\textbf{a}_1 & \textbf{a}_2 \\
| & |  \\
\end{bmatrix}
}_\text{\gls{6DRR}}.
\end{equation}
The neural network is then trained to predict these two columns. For a mapping $f$ to recover the full \gls{3D} rotation matrix, \cite{zhou_continuity_2019} propose to normalize and orthogonalize the predicted two columns and estimate the last one with the cross product as:
\begin{equation}
\begin{aligned}
 f 
\left( 
\begin{bmatrix}
| & | \\
\textbf{a}_1 & \textbf{a}_2 \\
| & | \\
\end{bmatrix}   
\right)
&= 
\begin{bmatrix}
| & | & | \\
\textbf{b}_1 & \textbf{b}_2 & \textbf{b}_3 \\
| & | & | \\
\end{bmatrix}   
\\
&=
\begin{bmatrix}
| & | & | \\
N(\textbf{a}_1) & O(\textbf{a}_2,\textbf{b}_1) & \textbf{b}_1 \times\textbf{b}_2 \\
| & | & | \\
\end{bmatrix}   
\end{aligned}
\end{equation}
where $N(\textbf{a}) = \frac{\textbf{a}}{||\textbf{a}||}$ and 
$O(\textbf{a},\textbf{b}) = N(\textbf{a} - (\textbf{b} \cdot \textbf{a})\textbf{b})$. Note the repeated use of $N(\textbf{a}_1)$ as $\textbf{b}_1$ here. %With this mapping we can utilize the \gls{6DRR} space of \cite{zhou_continuity_2019} for training a neural network to predict upper and lower arm rotations.

We investigate prediction accuracy for all discussed position and rotation representations: elbow and wrist positions in polar coordinates (Polar) and Cartesian coordinates (XYZ) a well as upper and lower arm rotations in \gls{6DRR} and quaternions (Quat). %The complete set of investigated representations for prediction targets is thus: $\lbrace \text{Polar}, \text{XYZ}, \gls{6DRR}, \text{Quat} \rbrace$,
% This results in comparing the feedforward and \gls{LSTM} neural network architectures on four distinct targets each.

\textbf{Activation Function} Also the choice for the activation function of a network has an effect on performance. Normalization of our \gls{IMU}, pressure or arm length inputs is cumbersome because of likely outliers. For example, extreme movements, like hitting an obstacle, can cause large spikes in accelerometer data. Additionally, it is difficult to define a minimum or maximum arm length since possible values vary between body proportions, children and adults. 

To mitigate the possible impact of out-of-distribution observations, we opted to employ the \gls{SELU} activation function by \cite{klambauer_self-normalizing_2017}, which is reported to induce self-normalizing properties. It is estimated as
\begin{equation}
    \text{SELU}(x) = \lambda 
    \begin{cases}
    x & \text{if}  x > 0  \\
    \alpha e^x - \alpha & \text{if} x\leq0
    \end{cases},
\end{equation}
where \cite{klambauer_self-normalizing_2017} derived $\alpha$ as $1.6733$ and $\lambda$ as $1.0507$. As summarized by \cite{klambauer_self-normalizing_2017}, these values enable necessary properties of the \gls{SELU} activation to allow for self normalization by, firstly, having positive and negative values for controlling the mean. Secondly, by featuring regions where the slope approaches zero and regions where the slope is larger than one. %For the latter case, the value of $\lambda > 1$ ensures that the slope is larger than one for positive $x$. 
These regions allow to dampen the variance if it is too large or to increase the variance if it is too low.
%Finally, the \gls{SELU} activation function is a continuous curve, which allows for approaching an equilibrium between damping and increasing the variance. 
With these properties, \cite{klambauer_self-normalizing_2017} showed that there are upper and lower bounds on the variance, thereby making learning robust even under the presence of noise and perturbations.

\textbf{Other Hyperparameters} For both architectures all layers consist of 128 neurons. The feedforward network features five layers and the \gls{LSTM} architecture four \gls{LSTM} layers. Both networks are trained for 200 epochs with the Adam \cite{kingma_adam_2015} optimizer, a learning rate of 0.001 and a \gls{MAE} loss function. Early stopping is applied when the minimal loss does not improve for 10 epochs. 

\subsection{Multimodality and Prediction Uncertainty}

% If predicted arm poses jump from a position below the chest to one above the head, the teleoperated robot may move too quickly and cause damage to itself or its surroundings.
% %Ensuring consistent and reliable predictions is crucial when using our smartwatch approach for robot teleoperation.
Even after incorporating the constraint of known body direction, still, the same smartwatch sensor recordings may have multiple possible arm posture solutions. To address this issue, we integrate dropout layers into our network architecture and utilize them for generating multiple stochastic forward passes through the network \cite{gal_dropout_2016}, i.e., \gls{MC} dropout predictions.

More specifically, \gls{MC} dropout predictions involve keeping the dropout activated for predictions outside of the training process and repeating every prediction multiple times. %, e.g., more than 100 repetitions. 
Producing repeated outputs with dropout results in a distribution of predictions for the same input. The standard deviation of the distribution serves a measure of the prediction uncertainty~\cite{gal_dropout_2016}. Such a distribution allows us to identify cases where smartwatch sensor readings lead to multiple possible arm postures. In such instances, the distribution can become multi-modal, and we can detect and choose the most likely mode based on additional constraints, such as the safest trajectory for the robot.

%When the standard deviation is high, indicating high uncertainty, we can instruct the robot to move following a more cautious policy.

\subsection{Speech recognition}

To further expand the teleoperation capabilities of our smartwatch approach, we incorporate the streaming of microphone data. The recorded audio signal is transcribed into voice commands utilizing the Google Cloud speech-to-text service\footnote{\url{https://cloud.google.com/speech-to-text}}. This additional interface proves effective and detects commands even when the arm of the user is hanging down. We demonstrate the usability of the speech recognition interface in \Cref{sec:usability-demonstrations}.

\section{RESULTS}

This section discusses and compares overall prediction accuracy of trained models. Further, we relate our findings to reported results in previous related work.

\subsection{Predictive Model Accuracy}

\begin{figure}[t]
\centerline{\includegraphics[width=0.5\textwidth]{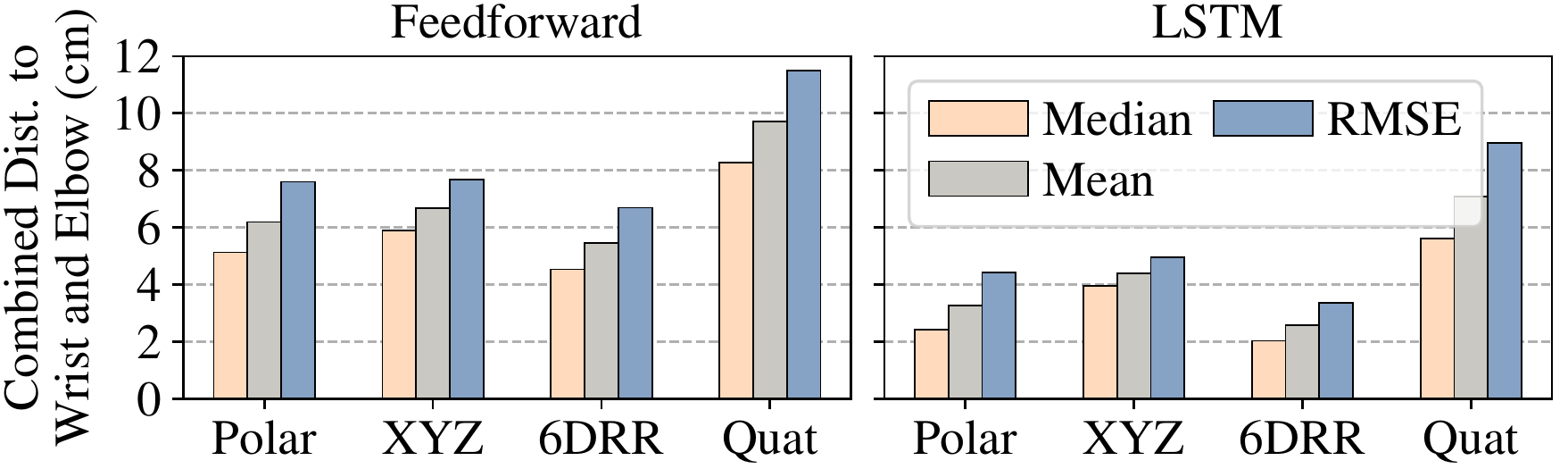}}
\caption{A comparison of prediction accuracy for combined wrist and elbow positions on test data. Both network architectures are trained to predict wrist and elbow positions in polar coordinates or Cartesian coordinates (XYZ) as well as upper and lower arm rotations as quaternions or \gls{6DRR}.}
\label{fig:model_comparison}
\end{figure}

\begin{figure}[t]
\centerline{\includegraphics[width=0.5\textwidth]{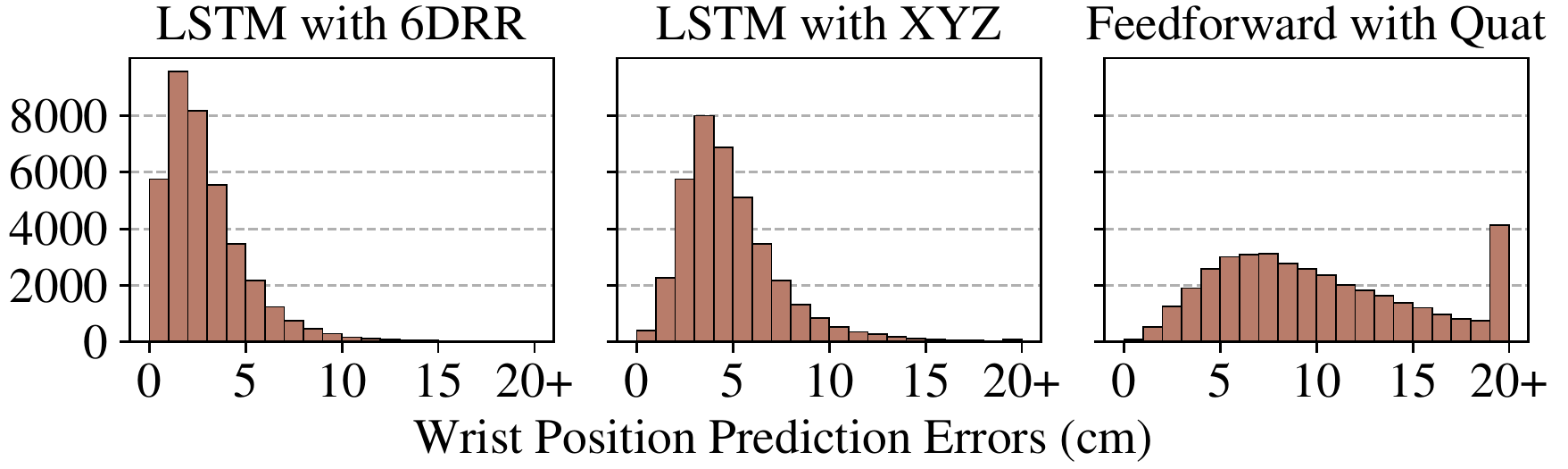}}
\caption{Error histograms of wrist position predictions of three distinct combinations of network architecture and prediction targets.}
\label{fig:histograms}
\end{figure}

\begin{figure}[t]
\centerline{\includegraphics[width=0.5\textwidth]{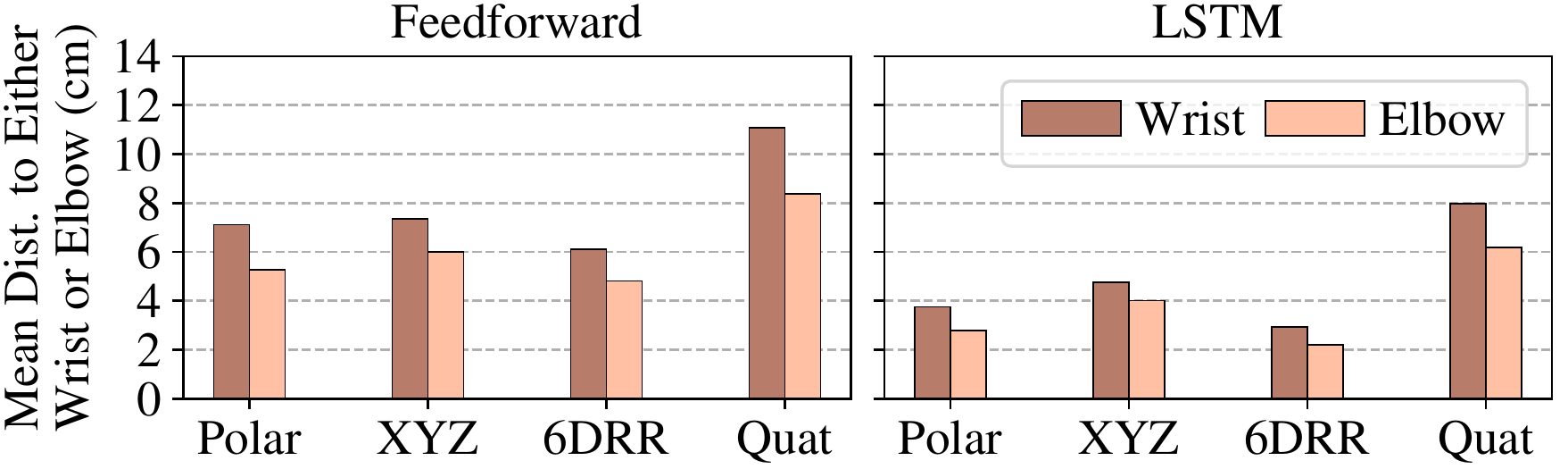}}
\caption{A comparison of mean prediction errors when focusing on either wrist or elbow positions.}
\label{fig:wrist_elbow_predictions}
\vspace{-\baselineskip}
\end{figure}

The feedforward and \gls{LSTM} network architectures are scrutinized by their prediction error on the test datasets via a 10-fold cross validation with each of our four introduced prediction targets Polar, XYZ, \gls{6DRR} and Quat. We derive the prediction error by calculating the combined distance from the predicted to the ground truth wrist and elbow positions divided by two. In \Cref{fig:model_comparison}, we compare the mean, the median error and \gls{RMSE} of those combined prediction errors.

Overall, the \gls{LSTM} models achieved lower errors than the feedforward models. This is an expected result given that arm movements are inherently a time series data set. The \gls{LSTM} architecture has an advantage since it maintains an internal state of previous rotations or accelerations thereby providing additional information to the prediction step.

Both network architectures achieved the lowest errors when trained on the \gls{6DRR} targets. This finding confirms that continuous rotation representations are more suitable training targets when compared to quaternions or Euler angles~\cite{zhou_continuity_2019}. Further, this finding validates our approach of optimizing the search space by utilizing the constraint that upper and lower arm lengths are constant. %Using a vanilla network with Cartesian XYZ prediction targets, any arbitrary position in space would have been a viable prediction. However,
Using the Polar, \gls{6DRR} and Quat targets, predictions are limited to the value ranges of the respective rotation spaces. The confirmed findings of \cite{zhou_continuity_2019} together with fixed arm lengths are plausible reasons for why \gls{6DRR} prediction targets achieve better performance.    

To investigate if reported average accuracy measurements hide extreme errors, \Cref{fig:histograms} depicts histograms of wrist position prediction errors. Each histogram summarizes the prediction errors for wrist positions of one fold during the conducted 10-fold cross validation. On the left in this comparison, the error distribution for the \gls{LSTM} with \gls{6DRR} combination shows the highest peak at the lowest error. The \gls{LSTM} with XYZ combination produces on average a higher prediction error, which is noticeable in a more right-shifted and wider error distribution. The combination of feedforward network with quaternion targets features a comparably flat error distribution and more than 4\,000 predictions with an error above 20\,cm. These observations coincide with our above findings that the \gls{LSTM} with \gls{6DRR} combination makes the most accurate predictions while the feedforward with quaternion combination is the least accurate. 

\Cref{fig:wrist_elbow_predictions} summarizes prediction errors for wrist and elbow positions independently. In general, it is observable that elbow predictions are more accurate than wrist predictions. This is plausible since the elbow has to lie on a sphere around the shoulder, while the wrist lies on a manifold defined by spheres around all possible elbow positions, allowing more room for error. Further, in case of the Polar, \gls{6DRR} and Quat targets, wrist positions are estimated by adding a vector with lower arm magnitude and with the predicted rotation onto the predicted elbow position. Thus, the error of the predicted elbow position potentially adds to the error of the predicted wrist position.

Altogether, the combination of \gls{LSTM} architecture and \gls{6DRR} targets outperforms other combinations with regards to prediction accuracy for wrist and elbow positions.

\subsection{Comparison to Related Work}

% We continue by with comparing these results to reported errors and methodology reported in related work. 
% To this end, they employ a hidden Markov model which maintains a probabilistic belief over the state of the system. 
The work of \cite{shen_i_2016} follows the the same objective as our paper, namely, the prediction of wrist and elbow positions from smartwatch data. Similar to our approach, they assume a fixed shoulder position and require the body facing direction to be known. In their evaluation they reported median errors of $9.2$\,cm for predicted wrist positions and $7.9$\,cm for elbow positions. Also \cite{wei_real-time_2021} used a recurrent neural network to predict wrist and elbow positions. They predict wrist and elbow positions in Cartesian coordinates and report an error of $7.2$\,cm and $7.1$\,cm for wrist elbow.

The \gls{LSTM} with \gls{6DRR} and coupled with \gls{MC} dropout predictions presented in our work is also suitable for real-time applications. Our Wear OS app allows to stream sensor data from the smartwatch to any reasonably well equipped system via UDP at $50$\,Hz. For example, with an $\text{Intel}^\text{\textregistered}$ $\text{Xeon}^\text{\textregistered}$ W-2125 CPU and a GeForce RTX 2080 Ti GPU we were able to make 150~\gls{MC} dropout predictions targets at a rate of $\sim40$\,Hz. Regarding the prediction accuracy, as reported in \Cref{fig:model_comparison}, our best performing model achieves a more than $4$\,cm reduction in median prediction errors compared to results reported by \cite{shen_i_2016, wei_real-time_2021}. Specifically, our model resulted in a median error of $2.33$\,cm for wrist position predictions and $1.61$\,cm for elbow predictions. 

% to predict wrist and elbow positions
% from
%  compared predictions to ground truth data from a motion capture system and 
Another related approach was proposed in \cite{joukov_human_2017}. However, their approach used two \gls{IMU}s; one \gls{IMU} on the lower arm and the second on the upper arm. In their real-world experiment they reported a \gls{RMSE} and standard deviation of $6.9\pm2.7$\,cm for wrist and $5.2\pm2.6$\,cm for elbow predictions. Their real-world experiment also required a short calibration procedure for their \gls{IMU}s based on the work of \cite{tedaldi_robust_2014}. 

As shown in \Cref{fig:model_comparison}, our \gls{LSTM} with \gls{6DRR} achieves a \mbox{$\sim40\%$} lower \gls{RMSE} on our motion capture data while relying only on a \emph{single} \gls{IMU}. Specifically, it predicts wrist positions with an \gls{RMSE} and standard deviation of $3.71\pm2.49$\,cm and elbow positions with an error of $2.99\pm2.19$\,cm on test data. In conclusion, our approach appears to result in a reduction of prediction error by at least $\sim40\%$ when  compared to previous works by \cite{shen_i_2016,joukov_human_2017,wei_real-time_2021} while it remains to be as real-time capable as the approach of \cite{shen_i_2016}. 

\section{USABILITY DEMONSTRATIONS}
\label{sec:usability-demonstrations}

We combine the increased accuracy of our approach with speech recognition through the microphone of the smartwatch and present two tasks which highlight the advantages of our approach. %Namely, the first use-case is an intervention task where we utilize the smartwatch to intervene in a predefined procedure of a robot and temporarily change its goal. The second use-case is a learning task where we leverage the smartwatch to swiftly collect a set of trajectories to teach a movement policy.

\subsection{Intervention Task}

\begin{wrapfigure}{r}{4.5cm}
\centerline{\includegraphics[width=0.2\textwidth]{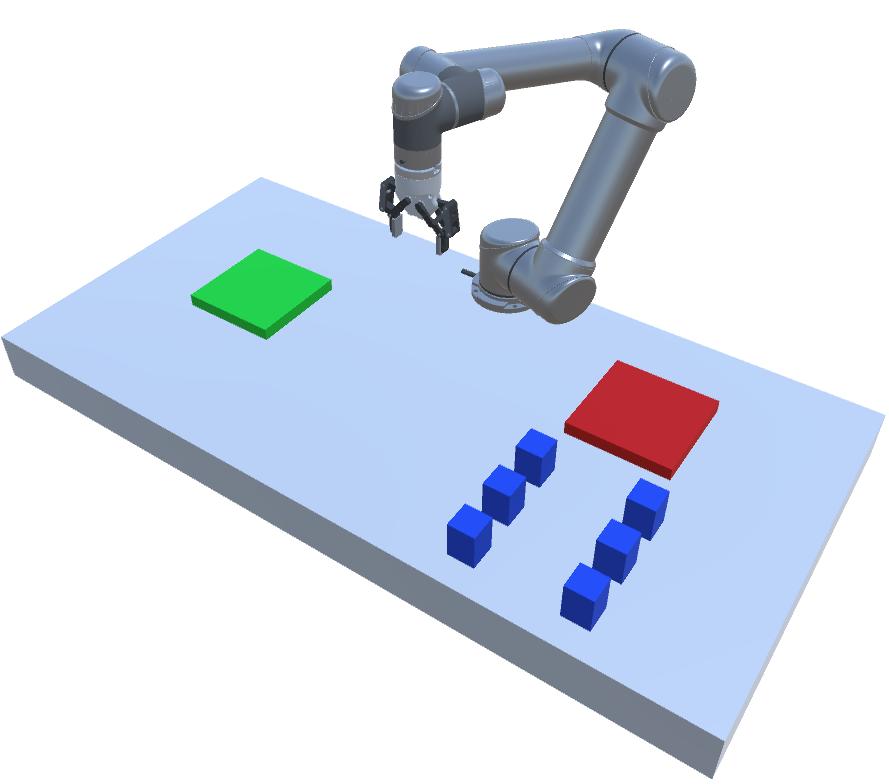}}
\caption{The intervention task: The robot picks the blue cubes one after the other and places them in the red area. The user utilizes the smartwatch to stop the robot mid-task and to move one cube to the green area instead.}
\label{fig:task_schmematic}
\end{wrapfigure}

This tasks demonstrates that the smartwatch allows for swift and intuitive human-robot-interaction at any time. %In this intervention task, a user is required to teleoperate a real UR5 robot. 
A schematic of the intervention task is depicted in \Cref{fig:task_schmematic}. The robot autonomously picks the blue cubes one-by-one and places them in the red area. The user can intervene at any time to place one of the cubes in the green area instead. Triggering an intervention is done via a voice command. Thereafter, the robot will mimic the human wrist motions.

\begin{figure}[t]
\centerline{\includegraphics[width=0.5\textwidth]{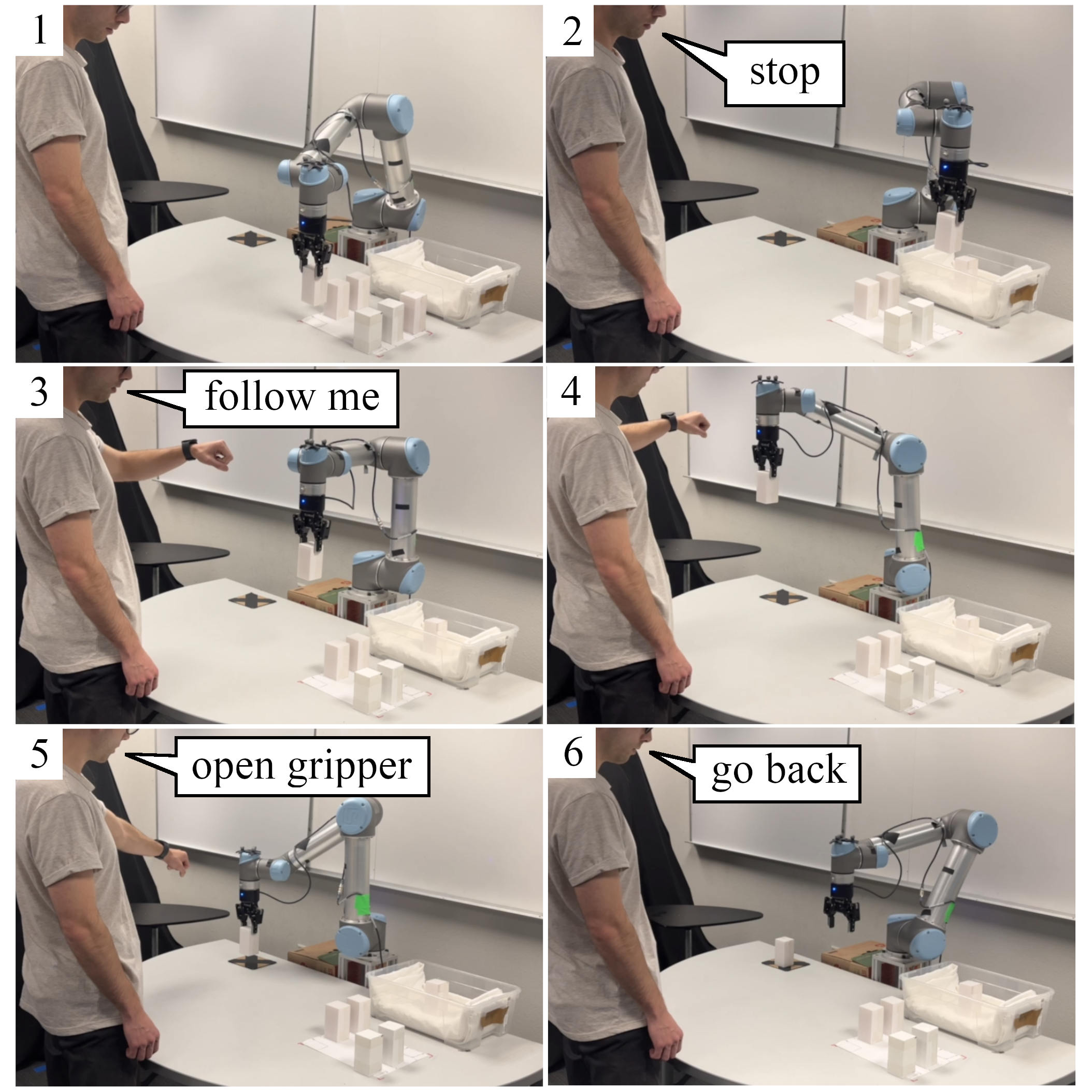}}
\caption{\textbf{Step~1:} The robot picks up cubes and puts them into the tray. \textbf{Step~2:} The user says ``stop''. The smartwatch recognizes the command and stops the robot mid-task. \textbf{Step~3:} The user raises their arm and says ``follow me''. The robot moves its end effector to match the wrist position. \textbf{Step~4:} The user guides the end effector to a marked position. \textbf{Step~5:} The user says ``open gripper'' and the robot drops the cube. \textbf{Step~6:} The user says ``go back'' and the robot returns to Step~1.}
\label{fig:task_steps}
\vspace{-\baselineskip}
\end{figure}

The entire procedure is subdivided into six steps, which are depicted and summarized in \Cref{fig:task_steps}. The tray in this real-world example is the red area from \Cref{fig:task_schmematic}, the black square on the left of the robot is the green area and the white cubes are arranged in front of the tray as the blue cubes in \Cref{fig:task_schmematic}.

%During step number one, the robot proceeds to place the cubes in the tray (red area in \Cref{fig:task_schmematic}). The second step is triggered when the user says ``stop''. For this, the user does not have to raise their arm, the smartwatch also recognizes the command when the arm is hanging down. It stops the robot mid-task. Then, the user raises their arm and says ``follow me''. Again, the smartwatch recognizes the command and sends its position to the robot. The robot moves to match end effector and smartwatch position. Now, the user moves their wrist to guide the end effector towards the marked area on the table (green area in \Cref{fig:task_schmematic}). Once the end effector hovers above the target area, the user tells the robot to drop the cube by saying ``open gripper''. As the sixth and final step, the user says ``go back'' and the robot proceeds to place the remaining cubes in the tray.

%and we measured the elapsed time between step two and step six as well as the distance between the marked target position and where the user placed the cube. 
Three users performed the task 10~times each. We measured the times from when the participant said ``stop'' until the ``open gripper'' command was received. The distances from the placed cube to the target marked positions were measured with retroreflective markers and the OptiTrack system which we used for our motion capture ground truth data.
%We attached four retroreflective markers around the target position and asked the participant to place the cube in the center of them. Additionally, we attached five retroreflective markers onto the cube which the participant places at the target position in step five. The center of these five markers were considered as the position of the cube. 
These distance and time measurements provided us with an estimate of how precisely the users could control the robot with the smartwatch and how quickly they could complete the task.

\begin{wraptable}{l}{4.6cm}
\begin{center}
\footnotesize
\caption{Intervention task results} \label{tab:intervention_task} 
\begin{tabular}{ 
c
S[table-format=2.1] @{${}\pm{}$} S[table-format=1.1]
S[table-format=1.2] @{${}\pm{}$} S[table-format=1.2]
}
\toprule 
Part. & \multicolumn{2}{c}{Time (s)} & \multicolumn{2}{c}{Dist. (cm)}\\
\midrule
1 & 19.5 & 2.6 & 1.87 & 0.62 \\
2 & 28.9 & 8.7 & 2.21 & 1.11 \\
3 & 20.1 & 4.7 & 2.19 & 1.07 \\
\midrule
All & 22.8 & 7.3 & 2.09 & 0.97 \\
\bottomrule 
\end{tabular}
\end{center}
\end{wraptable}
The results are summarized in \Cref{tab:intervention_task}. The average measured time from interrupting the robot until sending it back to its original task was $22.8\pm7.3$\,s. On average, all participants %(abbreviated as Part. in \Cref{tab:intervention_task}) 
placed their cubes within $2.09\pm0.97$\,cm from the target position. Every run was successful, which confirms that our smartwatch approach is a suitable tool for the designed intervention task. Further, considering the time and position error, these findings confirm the reported accuracy and real-time capability of our smartwatch approach. %The only requirement for the user to engage in this task was to perform the two-step calibration procedure before the task commenced. This interaction between robot and human is transferable to a multitude of other use-cases and shows that our smartwatch approach enables human-robot interaction anytime and anywhere. 

\subsection{Learning Task}

\begin{figure}[t]
\centerline{\includegraphics[width=0.439\textwidth]{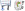}}
\caption{\textbf{Left:} A virtual concept of our learning task. We record six smartwatch trajectories for placing the blue cube onto the red locations. Then, we train a policy to generate new trajectories for placing the cube on the highlighted green positions. \textbf{Right:} The red training trajectories are recorded smartwatch data. The green trajectories were generated using our trained policy for the green target positions on the shelf.}
\label{fig:task_learn}
\vspace{-\baselineskip}
\end{figure}

The goal of the second task is to show an application to the problem of learning from demonstration~\cite{billard_robot_2008}. In particular, we learn a policy for placing a cube on a shelf, as depicted on the left in \Cref{fig:task_learn}. A human wearing a smartwatch demonstrates six training trajectories. The human holds the cube in heir hand at the start position and starts recording. Then, the human moves the cube in an arch to one of the six red marked positions on the shelf and repeats the procedure for the remaining goal positions. Since the human can demonstrate the trajectories without moving the robot, data collection is swift and uncomplicated. All training trajectories for this task were recorded within two minutes.

The smartwatch trajectories are depicted in red on the right in \Cref{fig:task_learn}. We then leverage these trajectories to train a movement policy using the \gls{GAIL}~\cite{ho_generative_2016} method. As a result, we obtain a movement policy for letting a robot place cubes at any target position on the shelf. To visualize the generalization capabilities of the resulting policy, four generated example trajectories are depicted in \Cref{fig:task_learn} on the right. They place the cube in-between the target positions, which are marked as green squares on the left in \Cref{fig:task_learn}.

This use-case demonstrates that the smartwatch can be leveraged to train new movement policies to a robot at any time by swiftly recording a set of demonstrations in the same environment. The smartwatch trajectories in this example were collected within two minutes and enable a robot to place a cube anywhere on a shelf given a target placement position. 

%We demonstrated the advantages of our ubiquitous smartwatch robot interface for robotics and showed that the most accurate trained model of this work achieves lower prediction errors for wrist and elbow position predictions than what related work of \cite{shen_i_2016,joukov_human_2017,wei_real-time_2021} reported for their real-world experiments.

\section{LIMITATIONS AND FUTURE WORK}
\label{sec:limt_and_future}
Our approach requires the completion of a two-step calibration procedure whenever there is a change in body orientation or location of the user. This is a limitation in comparison to the work of \cite{joukov_human_2017}. Their approach utilizes a second \gls{IMU} which allows users to move and rotate their hip and chest. This limitation can be addressed by adding a second \gls{IMU} to our smartwatch approach too. To maximize familiarity and ease-of-use, this work presents approaches to leverage the possibilities of the smartwatch to their full extend without adding additional devices. 
However, promising opportunities for future work can utilize the fact that a smartwatch is typically connected to a smartphone, which the user also wears on their body. The smartphone can serve as a second \gls{IMU} and %would allow the omission of the two-calibration procedure of our current approach. 
enable the tracking of arm movements while the user changes their body orientation or location. 

\begin{figure}[t]
\centerline{\includegraphics[width=0.5\textwidth]{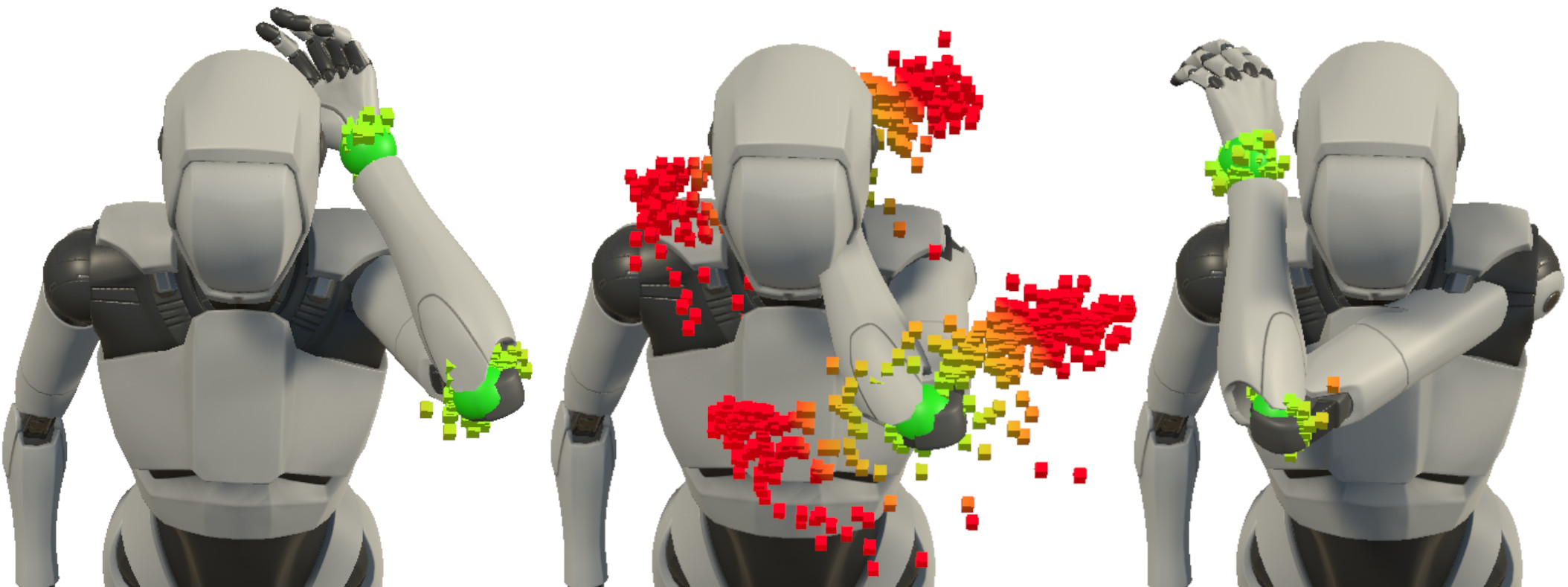}}
\caption{\textbf{Left:} The user wears the smartwatch on their left arm and holds the left hand next to their head. The smartwatch predicts the correct position. \textbf{Middle:} The user rotates their wrist back and forth while keeping the hand in the same position. This causes the predicted positions to alternate between positions left or right of the head. \textbf{Right:} The predicted wrist position is at the wrong side of the head.}
\label{fig:limitations}
\vspace{-\baselineskip}
\end{figure}

A further limitation is that fast wrist rotations or unergonomic arm motions affect the accuracy of our approach. \Cref{fig:limitations} illustrates an example where the user wears the smartwatch on their left arm and estimated arm postures are visualized with an avatar. The final predicted wrist and elbow positions are the mean of 300 individual \gls{MC} dropout forward passes. %The mean is highlighted as a green sphere and 
Individual predicted positions are marked as small cubes colored according to their distance to the mean. 

The user raised their hand to their ear and, as shown on the left in \Cref{fig:limitations}, the position was predicted correctly. Then, the user rotated their wrist back and forth while keeping their wrist position constant. The resulting unusual wrist angles and rapid movements caused predicted positions to alternate between the left and right side of the head. In the middle of \Cref{fig:limitations} it is observable that predictions manifested in bimodal distributions with their modes on the left an right side of the head. The mean, and therefore the predicted elbow and wrist positions, moved into the middle causing the arm of the avatar to go through its head. 

The detection and handling of such scenarios shapes promising opportunities for future work. The distributions obtained through the \gls{MC} dropout predictions allow to detect such scenarios and to dynamically adjust estimated joint positions. If a multimodal distribution occurs, we can consult additional cost functions, i.e., distance to previous positions or risk for the teleoperated robot. It will also be possible to determine the most likely arm posture by consulting additional predictive models, which were trained on different inputs. Having a measure of uncertainty and a distribution of possible solutions is a promising base to improve prediction accuracy in the future.

\section{CONCLUSIONS}

 %The task of estimating arm poses from smartwatch data poses a challenge as it can lead to multiple possible solutions for thesame smartwatch observations.
This work presents a solution to the problem of estimating human arm poses from a single smartwatch. We propose a simple yet effective two-step calibration procedure to mitigate variability in sensor data and to leverage information about the forward-facing direction of the user. This allows us to devise an optimized model architecture, which achieves a $\sim40\%$ reduction in prediction error compared to results reported in previous works. Furthermore, our approach generates a distribution of posture predictions, which allows to estimate a measure of uncertainty and to select the best solution from several options in cases of multimodal distributions. By combining arm posture estimations with speech recognition we turn the smartwatch into a ubiquitous, low-cost and versatile robot control interface. 

%This work demonstrates in an intervention task that three users successfully used the smartwatch to swiftly intervene in robot behavior and to temporarily adjust their goal. Further, we demonstrate in a learning task that users can collect trajectories with ease to train new behavior to a robot at any time.

\bibliographystyle{IEEEtran}
\scriptsize{
\bibliography{predictive_biomechanics}
}

\end{document}